\documentclass[11pt,a4paper]{article}
\usepackage[hyperref]{acl2020}
\usepackage{times}
\usepackage{latexsym}
\usepackage{amssymb}
\usepackage{mfirstuc}

\usepackage[normalem]{ulem}

\usepackage{microtype}
\usepackage{stmaryrd}
\aclfinalcopy % Uncomment this line for the final submission
\usepackage{graphicx}
\graphicspath{{figs/}}
\usepackage{csquotes}

\title{Some Grammatical Errors are Frequent, Others are Important}

\author{ 
  Leshem Choshen\\
  Department of Computer Science\\
  Hebrew University of Jerusalem\\
  {\tt\small leshem.choshen@mail.huji.ac.il}\\
  \And
  Ofir Shifman\\
  Department of Computer Science\\
  Hebrew University of Jerusalem\\
  {\tt\small ofir.shifman@mail.huji.ac.il }\\
  \AND 
  Omri Abend \\
  Department of Computer Science\\
  Hebrew University of Jerusalem\\
  {\tt\small omri.abend@mail.huji.ac.il}\\
}
\date{}

\begin{document}
\maketitle
\begin{abstract}
In Grammatical Error Correction, systems are evaluated by the number of errors they correct. However, no one has assessed whether all error types are equally important.
We provide and apply a method to quantify the importance of different grammatical error types to humans. We show that some rare errors are considered disturbing while other common ones are not. This affects possible directions to improve both systems and their evaluation.\footnote{All code and annotations are found in \url{https://github.com/borgr/GEC_BOTHER}}
\end{abstract}

%%%%%%%%%%%%%%%%%%%%%%%%%%%%%%
\section{Introduction}

Grammatical Error Correction (GEC) is the task of correcting erroneous human (mostly written; \citealp{Siddharth2015GrammaticalEC}) sentences. Predominantly, the sentences are writings of non-natives \citep{Wang2020ACS}. The use of this correction could be quite diverse, it could help communication, educate \citep{Obrien2015ConsciousnessRaisingEC, Tsai2020LinggleWriteAC}, evaluate \citep{gamon2013grammatical}, reduce language obstacles for learners \citep{wolfe2016grammatical} and more.\footnote{All code to replicate as well as the gathered data could be found in \url{https://github.com/borgr/GEC_BOTHER}}

In this work, we focus on the recipients of the grammatically erroneous text, rather than the writers. Doing so, we assess which types of errors are most important to correct. We follow a simplifying assumption that some errors inherently disrupt communication more than others, regardless of the sentence context. Under this assumption we ask native speakers to express their preference in partially erroneous sentences.

We manually annotate NUCLE \citep{dahlmeier2013building} erroneous sentences to find which ones are more crucial to correct (\S\ref{sec:annotation}). We then extrapolate the contribution of each type of error to the assessment of sentence correctness. Specifically, we train a linear predictor of the sentence score as a function of the amount of errors of each type (\S\ref{sec:type_pred}). From this we can not only know which error types' contribution is more important, without explicitly asking annotators about it, but also assess the contribution of each type to any typology of errors without further annotation.

Finally, computing the results on both the manual type system of NUCLE and automatic taxonomies, we find that some of the most frequent errors are of low importance and some infrequent ones are important, i.e., the errors which are most important to correct for humans and for current evaluation differ. Similarly, loss is implicitly weighted by frequency, but in this case frequency and importance differ. Thus, the emphasis in training is on the wrong types of errors.

%%%%%%%%%%%%%%%%%%%%%%%%%%%%%%%%%%%%%%
\section{Background}

Typologies of GEC error types date back to the early days of the field \cite{Dale2011HelpingOO}. Assuming each error stand by itself and is independent from other errors, each error could be given a class. Following this assumption manual annotations of typologies arrived with every dataset 
\citep{dahlmeier2013building,shatz2020refining} differing between them and between languages \citep{rozovskaya2019grammar,s21082658}.

Later, ERRANT proposed a method for automatically extracting errors from text and automatically annotating them with a set of rules \citep{bryant2017automatic}. This allowed to use the same annotation for any dataset in English. Lately, SErCl \citep{Choshen2020ClassifyingSE} proposed another typology, more fine-grained and based on syntax. It comes with an automatic extraction for most languages (depending on a part of speech tagger). SERRANT \citep{choshen2021serrant} combined the errors of ERRANT and SErCl to have a broader coverage, coming from SERRANT but use the meaningful rules for ERRANT categories. We do not give preferance to any of the methods and report results on each.

In most evaluation and literature, edit types are considered of equal importance, for example the $M^2$ \citep{dahlmeier2013building} scorer is based on errors corrected, regardless of their types. There are works however that show that models \citep{choshen2018inherent} and metrics \citep{choshen2018automatic} do not perform equally well on all error types. Specifically, they are better on closed class types where given that a valid correction was made, the reference is likely to correct in the same way and not perform another valid correction. Frequent types are also better addressed by learnt models, understandably. An exception to the above is \citet{gotou2020taking} that focuses on the most difficult types to correct. This is close in spirit to our work and valuable in itself. 

Knowing what is difficult to correct, as they suggest has merits. This knowledge may allow building a curriculum and highlight model failures. Still, we see our question as a more central one to the field, one that may shape the focus of future contributions for both models and evaluation. Difficulty to learn may change with technology, but what is perceived important to pursue will not. We propose an ideal for GEC to pursue and a way to measure it. 

Another work that is similar to ours in spirit is \citet{Tetreault2017JFLEGAF}, proposing to follow fluency rather that correct errors. In a sense, the most important errors to correct are those that most improve fluency of a text.

%%%%%%%%%%%%%%%%%%%%%%%%%%%%%%%%%%%%

\section{Annotation}\label{sec:annotation}
To get a reliable ranking of error importance we follow previous works' methodology. First, we do not ask annotators about grammaticality, as grammar in non-professionals is implicit and often even judged unimportant \citep{loewen2009second}. Instead, we ask annotators the extent to which a text is bothersome, following \citet{wolfe2016grammatical, Graham2015CanMT}. They found that impolite messages bothered job interviewers and to a lower extent so did ungrammatical writing. However, impolite texts were undeservedly judged ungrammatical, showing judges mix between the two.
%Following those evidence we chose not to rely on the notion of grammaticality.

We ask crowd annotators to directly assess the extent to which sentences need correction. We adapt the methodology of \citet{Graham2016IsAT} for assessing fluency of a text to assess instead how bothering a text is. Specifically, annotators were asked to move a slide to indicate how much they agree with the following: "The English mistakes in the following text bother me
(1 = it doesn’t bother me at all, 100 = it really bothers me)". All other details follow the original work.

We note that while we choose to follow common wording, other wordings may be acceptable and might even have slightly different results. For example, framing the question in terms of the context in which the sentence is written may produce different results. A sentence may be harshly judged in an academic writing but not in an email.

Every batch of sentences sent to the crowd contained 100 sentences ensuring that each annotator would produce at least 100 annotations. Only annotators from the United States with high (95\%>) acceptance rate and that reported they were English natives were accepted. This is to reduce noise due to faulty judgments and disagreements due to different countries of origin (e.g., native Australian citizens). Annotators were given 0.5\$ per batch, \footnote{The payment is not high, but by personal communication with the authors of Direct Assessment, high payment lures fraudulent annotators. Moreover, annotating the whole of NUCLE took less than two days, indicating that the payment was not deemed as low by the crowd annotators.} and their answers were normalized to follow a standard normal distribution (henceforth Z-score).

To allow filtering the data, each batch contains 3 types of sentences. 15 unique sentences which contain no mistakes. 70 unique sentences with at least one error. 15 sentences which were sampled from a a pre-sampled set of 400 sentences. The latter were repeatedly shown in different batches. The choice of 400 sentences was made to make sure a single annotator would not often see the same sentences and that we will have enough repetitions for each of the 400 to find outlying annotators. 

\subsection{Dataset}

We chose to annotate NUCLE \citep{dahlmeier2013building} containing about 59K sentences. Out of which we separated sentences with and without errors to two groups. Additionally, we filtered out sentences with less than 7 words, or ones that contained one of the strings: http, \&, [, ], *, ", ; to reduce non-English sentences. We also normalized spaces, deleting spaces after ) or before (, !, \%, ., \$, / and a comma (,).

We sent 58K sentences for annotation, which roughly corresponds to annotating each sentence with errors twice, plus multiple annotations of the 400 repetitive sets and about 8.7K annotations for grammatical sentences.

\subsection{Filtering}
An important aspect when asking for direct assessment from crowdworkers is to filter low quality annotations. We proceed to discuss this procedure.

\begin{figure}[tbp]
\includegraphics[width=8cm]{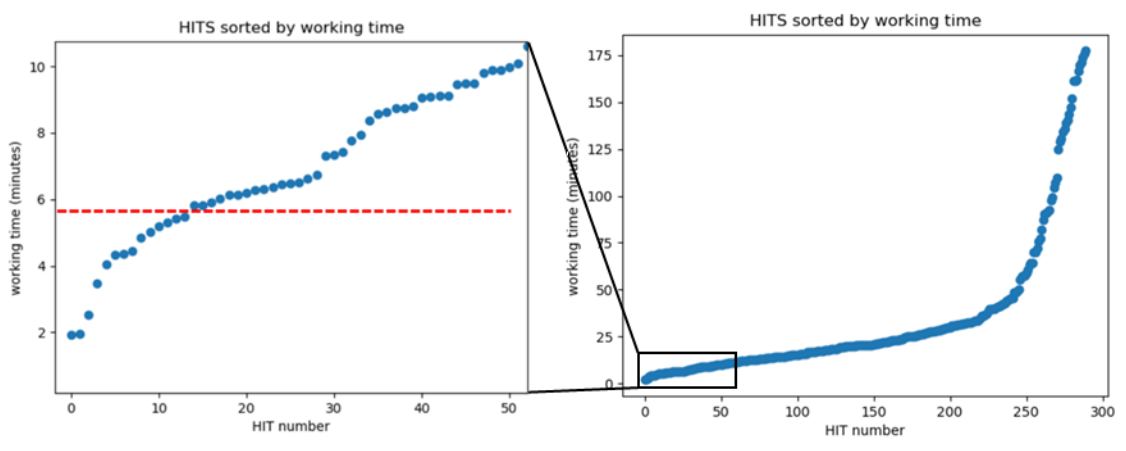}
\caption{Right: Working time per batch for one pass over NUCLE. Left: the tail of the distribution and in red the threshold below which annotators were filtered}
\label{fig:times}
% \vspace{-0.3cm}
\end{figure}

Annotators that took less than 350 seconds for 100 sentences were removed. Removing about 5\% of annotators (see Figure~\ref{fig:times}). This is expected to remove annotators who did not pay attention or mistakenly skipped a large number of sentences.

Among the remaining annotators, we made sure each judged the grammatical sentences to be better than the erroneous ones. Under the hypothesis that ungrammatical sentences had a lower score, we made a t-test for each annotator. If the grammatical sentences did not have a significantly higher average sentence score than the ungrammatical ($p<0.05$), we filtered out all the annotations made by the annotator. Overall about 2\% of annotators were filtered in this method.

Last, we compared the Pearson correlation between each annotator's Z-scores and the rest's on the repeating sentences. Following \citet{graham2015accurate}, correlation only took into account sentences with at least 15 responses as the average is noisy. Annotators with strong negative correlations ($>-0.4$) were filtered out. Overall, these procedures filtered about 10\% of the annotators.

Furthermore, we found most annotators filtered in the previous stages had negative correlation, which validates this methodology, as the different filtering methods agree. Raising the bars of either P or minimum time had diminishing gains in terms of finding negative correlation annotators.

While the annotations still contain noise, trying to filter out more with harsher thresholds produced similar results (See \S\ref{sec:results}) with more variance (due to less data). This suggests that the results are robust to this filtering and are reliable in that sense.

% \begin{figure}[tbp]
% % \includegraphics[width=8cm]{correlation.png}
% \caption{Correlation between each annotator and the rest sorted by correlation.}
% \label{fig:times}
% % \vspace{-0.3cm}
% \end{figure}

\section{Score per Type}\label{sec:type_pred}

As mentioned above, annotations are done on a sentence level. While this means we need to extrapolate which type of error is more important, it also allows us to do it for different error annotation schemes.

We experiment with both the manual annotated error types in the NUCLE corpus and automatic error types. Specifically, we analyse both automatic error types of ERRANT \citep{errant} and SErCL \citep{Choshen2020ClassifyingSE}. We do not analyze SERRANT \citep{choshen2021serrant} as it is based on the two latter and is hence quite similar.

Given the sentence scores we train a linear classifier with the error types count as features. For each sentence, we extract the number of times each type of error was found in it. We then train the linear regression to predict the annotation score based on these features.
%Ofir: I'm pretty sure I also check it with binary vectors - meaning that we only checked whether a sentence contains some error type (and didn't count multiple mistakes of same type, results were very similar - basically the same)

The output weights can be understood as the contribution of each type to the sentence annoyance levels. Note that in doing so, we assume a linear contribution of types. Namely, that when multiple types appear or a single type appears more than once, their contribution is additive. Future work may consider more complex extrapolations with softer assumptions.

Because the actual weights are hard to interpret, we focus on the ranks of each phenomena. In other words, we look to see who got the largest weight, the second largest and so on, rather than the actual distribution of weights that were assigned (we report those for completeness in App.~\ref{ap:weights}).

We extrapolate for each NUCLE type, for SErCl's most frequent types, for ERRANT's types and for ERRANT's types without sub-categorization to replacement additions and deletions.

\section{Results} \label{sec:results}

\begin{figure}[tbp]
\includegraphics[width=8cm]{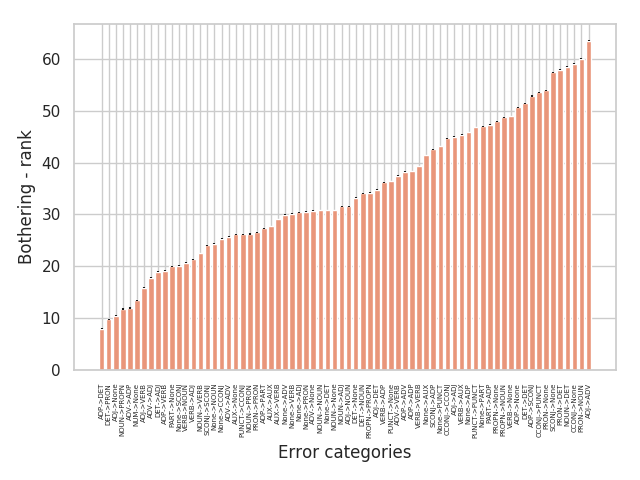}
\caption{Importance ranks of each SErCl type. Std in error bars.}
\label{fig:sercl_rank}
% \vspace{-0.3cm}
\end{figure}
\begin{figure}[tbp]
\includegraphics[width=8cm]{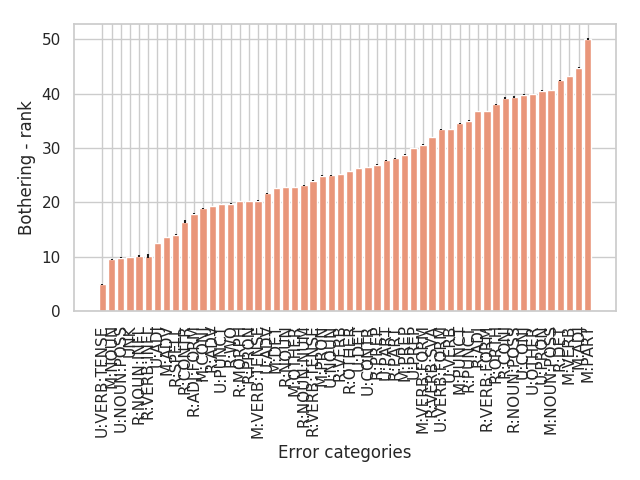}
\caption{Importance ranks of each ERRANT type. Std in error bars. (fine-grained ERRANT types in Appendix \ref{ap:weights}.)
\label{fig:ERRANT_rank}}
% \vspace{-0.3cm}
\end{figure}

\begin{figure}[tbp]
\includegraphics[width=8cm]{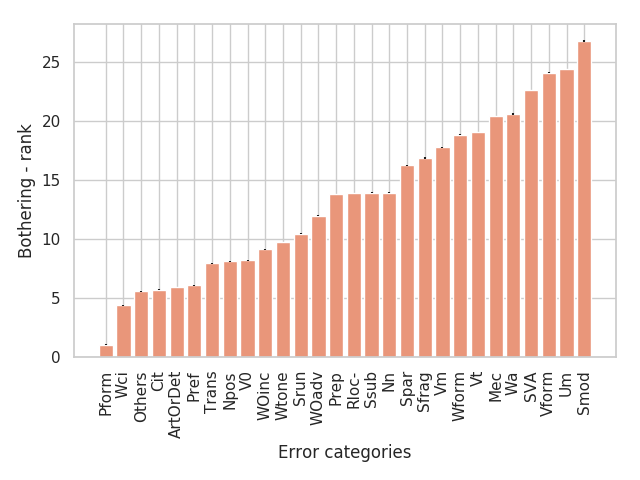}
\caption{Importance ranks of each NUCLE type. Std in error bars.}
\label{fig:NUCLE_rank}
% \vspace{-0.3cm}
\end{figure}

We present the ranking for SErCl in Fig. \ref{fig:sercl_rank}, for ERRANT in Fig. \ref{fig:ERRANT_rank} (Fine grained with insertion deletion and modification in App.~\ref{ap:weights}) and for NUCLE in Fig. \ref{fig:NUCLE_rank}. We also report the actual weight in appendix \ref{ap:weights} and note that those are more variable and harder to reason about. 

We see that despite the large sample there is still variance. Thus, some error types are not significantly harder than others. Still, which errors are easy, medium or hard is clear.

We find that, across the typologies, verb inflection and verb errors in general are among most bothering errors. So are orthography errors, unnecessarily added tokens, wrong determiner and other errors.

On the other side of the spectrum we can find missing tokens, inflection, morphology and others. Several errors related to determiners are also low ranking.

\section{Discussion and Conclusion}

Most metrics disregard the error type, at least in principal \citep[][In practice errors are unintentionally weighted differently, but not by design;]{choshen2018automatic}. This has been criticized and difficulty of correction was suggested to address it \citep{gotou2020taking}. Our results show that not only some errors are more important to correct than others, those are not determined by frequency in the data nor in the difficulty to correct. Determiners are extremely common and a closed class \citep{choshen2018inherent}, making them more important to correct to gain high scores in metrics, but those errors are not considered very important by humans. Similarly, orthographic errors are very easy to correct, but they are considered very annoying and important to correct.

We also performed initial studies with weighting training spans by giving each token its weight by the importance of the error (non-error tokens weight is constant). Unsurprisingly, the network improves over the relevant errors more than on others or the baseline, although not by a large margin.

\section{Acknowledgments}
We thank Dan Malkin for the experiments with weighted gradients.%\oa{what do you thank him for exactly? what did he do?}\lc{he did the experiments in the paragraph just before this one:We also performed initial studies with weighting training spans[...]}

\bibliography{acl2020}
\bibliographystyle{acl_natbib}

\clearpage
\appendix
\section{Additional Graphs}\label{ap:weights}
We present here the fine-grained ERRANT labels and the linear regression weights with their std. Note that negative score does not necessarily means that this type is considered positive by annotators, as there is a baseline too (so it might only be less severe than other errors).
\begin{figure}[tbp]
\includegraphics[width=8cm]{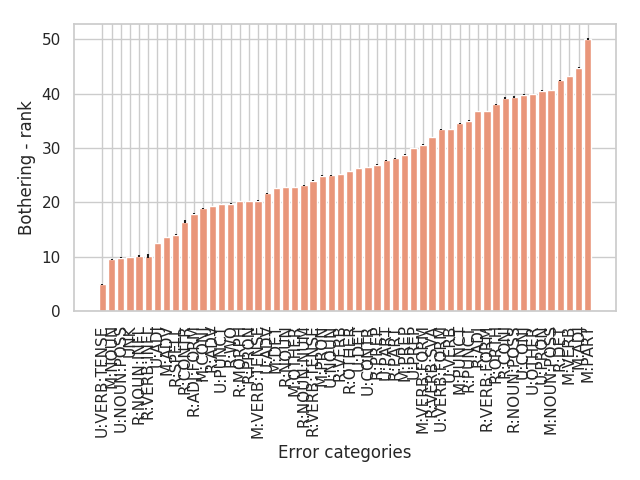}
\caption{Importance ranks of each coarse-grained ERRANT types. Std in error bars.}
\label{fig:coarseERRANT_rank}
% \vspace{-0.3cm}
\end{figure}
\begin{figure}[tbp]
\includegraphics[width=8cm]{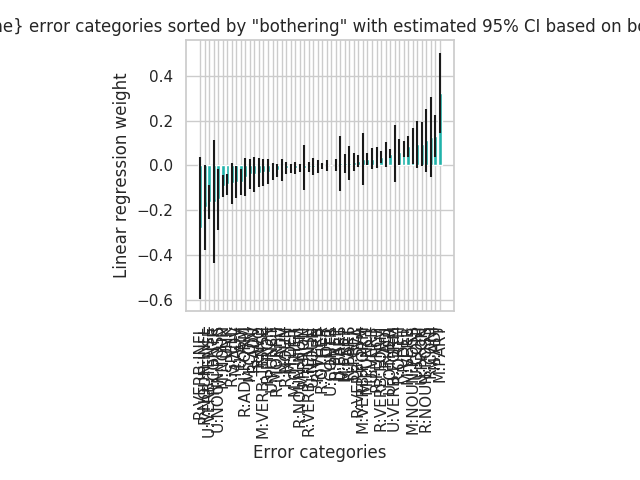}
\caption{Importance weights of each fine-grained ERRANT types. Std in error bars.}
\label{fig:fineERRANT_rank}
% \vspace{-0.3cm}
\end{figure}

\begin{figure}[tbp]
\includegraphics[width=8cm]{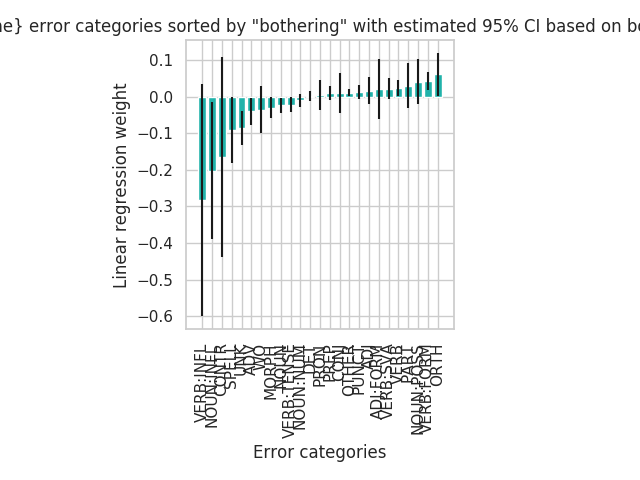}
\caption{Importance weights of each coarse-grained ERRANT types. Std in error bars.}
\label{fig:coarseERRANT_weight}
% \vspace{-0.3cm}
\end{figure}

\begin{figure}[tbp]
\includegraphics[width=8cm]{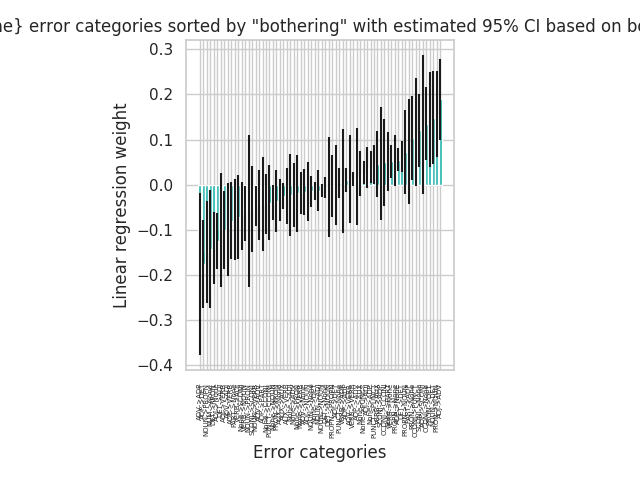}
\caption{Importance weights of each SErCl type. Std in error bars.}
\label{fig:SERCL_weight}
% \vspace{-0.3cm}
\end{figure}

\begin{figure}[tbp]
\includegraphics[width=8cm]{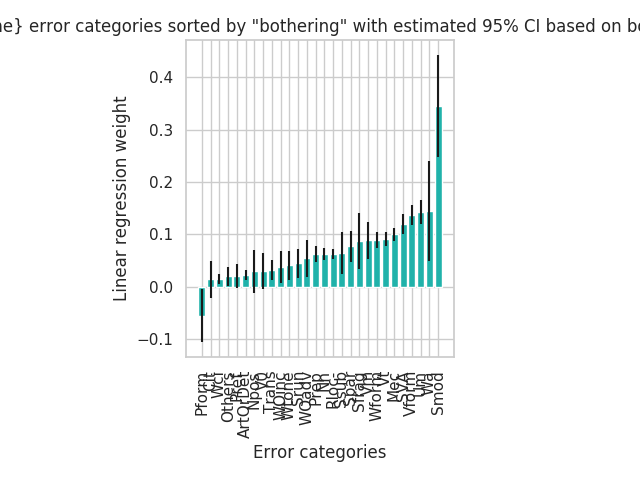}
\caption{Importance weights of each NUCLE type. Std in error bars.}
\label{fig:NUCLE_weight}
% \vspace{-0.3cm}
\end{figure}
\end{document}